\documentclass{article}
\usepackage{spconf,amsmath,graphicx}
\usepackage{amssymb}
\usepackage{amsfonts}
\usepackage{algorithm,algorithmic,setspace}
\usepackage{epsfig}

\def\BE{\vspace{-2.0mm}\begin{equation}}
\def\EE{\vspace{-1.4mm}\end{equation}}
\def\BEA{\vspace{-2.0mm}\begin{eqnarray}}
\def\EEA{\vspace{-1.0mm}\end{eqnarray}}

\newcommand{\alg}[1]{Algorithm \ref{alg:#1}}
\newcommand{\eqn}[1]{Eqn.~\ref{eqn:#1}}
\newcommand{\fig}[1]{Fig.~\ref{fig:#1}}
\newcommand{\tab}[1]{Table~\ref{tab:#1}}

\def\etal{{\textit{et~al.~}}}
\def\units{{\text{units of }}}

\title{ADADELTA: An Adaptive Learning Rate Method}

\name{Matthew D. Zeiler$^{\textrm{1,2}}$\sthanks{This work was done while
    Matthew D. Zeiler was an intern at Google.}}
\address{$^{\textrm{1}}$Google Inc., USA \hspace{1cm} $^{\textrm{2}}$New York University, USA}

\begin{document}
\maketitle
\begin{abstract}
  We present a novel per-dimension learning rate method for gradient descent called ADADELTA.
  The method dynamically adapts over time using only first order
  information and has minimal computational overhead beyond vanilla stochastic gradient
  descent. The method requires no manual tuning of a learning rate and appears robust to noisy
  gradient information, different model architecture choices, various data modalities and selection of hyperparameters.
  We show promising results compared to other methods on the MNIST digit classification task
  using a single machine and  on a large scale voice dataset in a distributed cluster environment.
\end{abstract}


\section{Introduction} \label{sec:intro}

The aim of many machine learning methods is to update a set of parameters $x$ in order to
optimize an objective function $f(x)$. This often involves some iterative procedure which applies
changes to the parameters, $\Delta x$ at each iteration of the algorithm. Denoting the parameters at
the $t$-th iteration as $x_t$, this simple update rule becomes:
\BE
x_{t+1} = x_t + \Delta x_t
\label{eqn:update}
\EE
In this paper we consider gradient descent algorithms which attempt to optimize the objective
function by following the steepest descent direction given by the negative of the gradient
$g_t$. This general approach can be applied to update any parameters for which a derivative can be
obtained:
\BE
\Delta x_t = - \eta g_t
\label{eqn:change}
\EE
where $g_t$ is the gradient of the parameters at the $t$-th iteration
$\frac{\partial f(x_t)}{\partial x_t}$
and $\eta$ is a learning rate which controls how large of a step to take in
the direction of the negative gradient. Following this negative gradient for each new sample or
batch of samples chosen from the dataset gives a local estimate of which direction minimizes the
cost and is referred to as stochastic gradient descent (SGD) \cite{robbins51}. While often simple to
derive the gradients for each parameter analytically, the gradient descent algorithm requires the
learning rate hyperparameter to be chosen.

Setting the learning rate typically involves a tuning procedure in which the highest possible
learning rate is chosen by hand. Choosing higher than this rate can cause the system to diverge
in terms of the objective function, and choosing this rate too low results in slow learning.
Determining a good learning rate becomes more of an art than science for many problems.

This work attempts to alleviate the task of choosing a learning rate by introducing a new dynamic
learning rate that is computed on a per-dimension basis using only first order
information. This requires a trivial amount of extra computation per iteration over gradient
descent. Additionally, while there are some hyper parameters in this method, we has found
their selection to not drastically alter the results. The benefits of this approach are as follows:
\begin{itemize}
\item no manual setting of a learning rate.\vspace{-2mm}
\item insensitive to hyperparameters.\vspace{-2mm}
\item separate dynamic learning rate per-dimension.\vspace{-2mm}
\item minimal computation over gradient descent.\vspace{-2mm}
\item robust to large gradients, noise and architecture choice.\vspace{-2mm}
\item applicable in both local or distributed environments. \vspace{-2mm}
\end{itemize}

\section{Related Work} \label{sec:related}
There are many modifications to the gradient descent algorithm. The most powerful such
modification is Newton's method which requires second order derivatives of the cost function:
\BE
\Delta x_t = H_t^{-1} g_t
\EE
where $H_t^{-1}$ is the inverse of the Hessian matrix of second derivatives computed at iteration $t$. This
determines the optimal step size to take for quadratic problems, but unfortunately is prohibitive to
compute in practice for large models.
Therefore, many additional approaches have been proposed to either improve the use of first order
information or to approximate the second order information.

\subsection{Learning Rate Annealing}
There have been several attempts to use heuristics for estimating a good learning rate at each
iteration of gradient descent. These either attempt to speed up learning when suitable or to slow
down learning near a local minima. Here we consider the latter.

When gradient descent nears a minima in the cost surface, the parameter values can oscillate back
and forth around the minima. One method to prevent this is to
slow down the parameter updates by decreasing the learning rate. This can be done manually when the
validation accuracy appears to plateau. Alternatively, learning rate schedules have been proposed
\cite{robbins51} to automatically anneal the learning rate based on how many epochs through the data
have been done. These approaches typically add additional hyperparameters to control how quickly the
learning rate decays.

\subsection{Per-Dimension First Order Methods}
The heuristic annealing procedure discussed above modifies a single global learning rate that applies to
all dimensions of the parameters. Since each dimension of the parameter vector can relate to the
overall cost in completely different ways, a per-dimension learning rate that can compensate for
these differences is often advantageous.

\subsubsection{Momentum}
One method of speeding up training per-dimension is the momentum method \cite{autoencoder-nn}. This
is perhaps
the simplest extension to SGD that has been successfully used for decades. The main idea behind
momentum is to accelerate progress along dimensions in which gradient consistently point in the same
direction and to slow progress along dimensions where the sign of the gradient continues to
change. This is done by keeping track of past parameter updates with an exponential decay:
\BE
\Delta x_t = \rho \Delta x_{t-1} - \eta g_t
\EE
where $\rho$ is a constant controlling the decay of the previous parameter updates. This gives a
nice intuitive improvement over SGD when optimizing difficult cost surfaces such as a long narrow
valley. The gradients along the valley, despite being much smaller than the gradients across the
valley, are typically in the same direction and thus the momentum term accumulates to speed up
progress. In SGD the progress along the valley would be slow since the gradient
magnitude is small and the fixed global learning rate shared by all dimensions cannot speed up
progress. Choosing a higher learning rate for SGD may help but the dimension across the valley
would then also make larger parameter updates which could lead to oscillations back as forth across
the valley. These oscillations are mitigated when using momentum because the sign of the gradient
changes and thus the momentum term damps down these updates to slow progress across the
valley. Again, this occurs per-dimension and therefore the progress along the valley is unaffected.

\subsubsection{ADAGRAD}
A recent first order method called ADAGRAD \cite{adagrad} has shown remarkably good results on large
scale learning tasks in a distributed environment \cite{gbrain-nips12}.
This method relies on only first order information but has some properties of second order
methods and annealing. The update rule for ADAGRAD is as follows:
\BE
\Delta x_t = -  \frac{\eta}{\sqrt{\sum_{\tau=1}^t g_{\tau}^2}} \; g_t
\label{eqn:ADAGRAD}
\EE
Here the denominator computes the $\ell2$ norm of all previous gradients on a per-dimension basis
and $\eta$ is a global learning rate shared by all dimensions.

While there is the hand tuned global learning rate, each dimension has its own dynamic rate. Since
this dynamic rate grows with the inverse of the gradient magnitudes, large gradients have smaller
learning rates and
small gradients have large learning rates. This has the nice property, as in second order methods,
that the progress along each dimension evens out over time. This is very beneficial for training
deep neural networks since the scale of the gradients in each layer is often different by several
orders of magnitude, so the optimal learning rate should take that into account. Additionally, this
accumulation of gradient in the denominator has the same effects as annealing, reducing the learning
rate over time.

Since the magnitudes of gradients are factored out in ADAGRAD, this method can be sensitive to
initial conditions of the parameters and the corresponding gradients. If the initial gradients are
large, the learning rates will be low for the remainder of training.
This can be combatted by increasing the global learning rate, making the ADAGRAD method sensitive to
the choice of learning rate. Also, due to the continual accumulation of
squared gradients in the denominator, the learning rate will continue to decrease throughout
training, eventually decreasing to zero and stopping training completely.
We created our ADADELTA method to overcome the sensitivity to the hyperparameter selection as well
as to avoid the continual decay of the learning rates.

\subsection{Methods Using Second Order Information}
Whereas the above methods only utilized gradient and function evaluations in order to optimize the
objective, second order methods such as Newton's method or quasi-Newtons methods make use of the
Hessian matrix or approximations to it. While this provides additional curvature information useful
for optimization, computing accurate second order information is often expensive.

Since computing the entire Hessian matrix of second derivatives is too
computationally expensive for large models, Becker and LecCun \cite{becker1988} proposed a diagonal
approximation to the Hessian. This diagonal approximation can be computed with one additional
forward and back-propagation through the model, effectively doubling the computation over
SGD. Once the diagonal of the Hessian is computed, $diag(H)$, the update rule becomes:
\BE
\Delta x_t = - \frac{1}{|\text{diag}(H_t)| + \mu} \; g_t
\EE
where the absolute value of this diagonal Hessian is used to ensure the negative gradient direction is always
followed and $\mu$ is a small constant to improve the conditioning of the Hessian for regions of
small curvature.

A recent method by Schaul \etal \cite{schaul2012} incorporating the diagonal Hessian with ADAGRAD-like terms has
been introduced to alleviate the need for hand specified learning rates. This method uses the
following update rule:
\BE
\Delta x_t = - \frac{1}{|\text{diag}(H_t)|} \frac{E[g_{t-w:t}]^2}{E[g_{t-w:t}^2]} \; g_t
\EE
where $E[g_{t-w:t}]$ is the expected value of the previous $w$ gradients and $E[g^2_{t-w:t}]$ is the
expected value of squared gradients over the same window $w$. Schaul \etal also introduce a
heuristic for this window size $w$ (see \cite{schaul2012} for more details).

\section{ADADELTA Method} \label{sec:method}
The idea presented in this paper was derived from ADAGRAD \cite{adagrad} in order to improve upon
the two main drawbacks of the method: 1) the continual decay of learning rates throughout training,
and 2) the need for a manually selected global learning rate. After
deriving our method we noticed several similarities to Schaul \etal \cite{schaul2012}, which will be
compared to below.

In the ADAGRAD method the denominator accumulates the squared gradients from each iteration starting
at the beginning of training. Since each term is positive, this accumulated sum continues to grow
throughout training, effectively shrinking the learning rate on each dimension. After many
iterations, this learning rate will become infinitesimally small.

\subsection{Idea 1: Accumulate Over Window}
Instead of accumulating the sum of squared gradients over all time, we restricted the window of
past gradients that are accumulated to be some fixed size $w$ (instead of size $t$ where $t$ is the
current iteration as in ADAGRAD). With this windowed accumulation the denominator of ADAGRAD cannot
accumulate to infinity and instead becomes a local estimate using recent gradients. This ensures
that learning continues to make progress even after many iterations of updates have been done.

Since storing $w$ previous squared gradients is inefficient, our methods implements this
accumulation as an exponentially decaying average of the squared gradients. Assume at time $t$ this
running average is $E[g^2]_t$ then we compute:
\BE
E[g^2]_t = \rho \; E[g^2]_{t-1} + (1-\rho) \; g_t^2
\EE
where $\rho$ is a decay constant similar to that used in the momentum method. Since we require the
square root of this quantity in the parameter updates, this effectively becomes the RMS of previous
squared gradients up to time $t$:
\BE
\text{RMS}[g]_t = \sqrt{E[g^2]_t + \epsilon}
\EE
where a constant $\epsilon$ is added to better condition the denominator as in \cite{becker1988}. The resulting parameter
update is then:
\BE
\Delta x_t = -  \frac{\eta}{\text{RMS}[g]_t} \; g_t
\label{eqn:idea1}
\EE

\subsection{Idea 2: Correct Units with Hessian Approximation}
When considering the parameter updates, $\Delta x$, being applied to $x$, the units should
match. That is, if the parameter had some hypothetical units, the changes to the parameter should
be changes in those units as well.
When considering SGD, Momentum, or ADAGRAD, we can see that this is not the case.
The units in SGD and Momentum relate to the gradient, not the parameter:
\BE
\units \Delta x \propto \units g \propto \frac{\partial f}{\partial x} \propto \frac{1}{\units \; x}
\EE
assuming the cost function, $f$, is unitless. ADAGRAD also does not have correct units since the
update involves ratios of gradient quantities, hence the update is unitless.

In contrast, second order methods such as Newton's method that use Hessian information or an
approximation to the Hessian do have the correct units for the parameter updates:
\BEA
\Delta x \propto H^{-1} g \propto
\frac{\frac{\partial f}{\partial x}}{\frac{\partial^2 f}{\partial x^2}}
\propto \units x
\EEA

\begin{algorithm}[t!]
\small
\begin{algorithmic}[1]
\REQUIRE Decay rate $\rho$,  Constant $\epsilon$
\REQUIRE Initial parameter $x_1$
\STATE Initialize accumulation variables $E[g^2]_0 = 0$, $E[\Delta x^2]_0 = 0$
\FORC{$t=1:T$}{ \%\% Loop over \# of updates}
\STATE Compute Gradient: $g_t$
\STATE Accumulate  Gradient: $E[g^2]_t = \rho E[g^2]_{t-1} + (1-\rho) g_t^2$
\STATE Compute Update: $\Delta x_t = - \frac{\text{RMS}[\Delta x]_{t-1}}{\text{RMS}[g]_t} \; g_t$
\STATE Accumulate Updates:  $E[\Delta x^2]_t = \rho E[\Delta x^2]_{t-1} + (1-\rho)
\Delta x_t^2$
\STATE Apply Update: $x_{t+1} = x_t + \Delta x_t$
\ENDFOR
\end{algorithmic}
\small
\caption{Computing ADADELTA update at time $t$}
\small
\label{alg:am}
\end{algorithm}

Noticing this mismatch of units we considered terms to add to \eqn{idea1} in order for the
units of the update to match the units of the parameters.
Since second order methods are correct, we rearrange Newton's method (assuming a diagonal Hessian) for the inverse
of the second derivative to determine the quantities involved:
\BEA
\Delta x = \frac{\frac{\partial f}{\partial x}}{\frac{\partial^2 f}{\partial x^2}}
\Rightarrow
\frac{1}{\frac{\partial^2 f}{\partial x^2}} =  \frac{\Delta x}{\frac{\partial f}{\partial x}}
\label{eqn:newton}
\EEA

Since the RMS of the previous gradients is already represented in the denominator in \eqn{idea1} we
considered a measure of the $\Delta x$ quantity in the numerator. $\Delta x_t$ for the current time
step is not known, so we assume the curvature is locally smooth and approximate $\Delta x_t$
by compute the exponentially decaying RMS over a window of size $w$ of previous $\Delta x$ to give
the ADADELTA method:
\BE
\Delta x_t = - \frac{\text{RMS}[\Delta x]_{t-1}}{\text{RMS}[g]_{t}} \; g_t
\label{eqn:adadelta}
\EE
where the same constant $\epsilon$ is added to the numerator RMS as well. This constant serves the
purpose both to start off the first iteration where $\Delta x_0 = 0$ and to ensure progress
continues to be made even if previous updates become small.

This derivation made the assumption of diagonal curvature so that the second derivatives
could easily be rearranged. Furthermore, this is an approximation to the diagonal Hessian using only
$\text{RMS}$ measures of $g$ and $\Delta x$. This approximation is always positive
as in Becker and LeCun \cite{becker1988}, ensuring the update direction follows the
negative gradient at each step.

In \eqn{adadelta} the $\text{RMS}[\Delta x]_{t-1}$ quantity lags behind the denominator by 1 time
step, due to the recurrence relationship for $\Delta x_t$. An interesting side effect of this is
that the system is robust to large sudden gradients which act to increase the denominator, reducing
the effective learning rate at the current time step, before the numerator can react.

The method in \eqn{adadelta} uses only first order information and has some properties from each of
the discussed methods. The negative gradient direction for the current iteration $-g_t$ is always
followed as in SGD. The numerator acts as an acceleration term, accumulating previous gradients over
a window of time as in momentum. The denominator is related to ADAGRAD in that the squared gradient
information per-dimension helps to even out the progress made in each dimension, but is computed
over a window to ensure progress is made later in training. Finally, the method relates to Schaul
\etal's in that some approximation to the Hessian is made, but instead costs only one gradient
computation per iteration by leveraging information from past updates. For the complete algorithm
details see \alg{am}.

\section{Experiments} \label{sec:experiments}

We evaluate our method on two tasks using several different neural network architectures. We train
the neural networks using SGD, Momentum, ADAGRAD, and ADADELTA in a supervised fashion to minimize
the cross entropy objective between the network output and ground truth labels.
Comparisons are done both on a local computer and in a distributed compute cluster.

\subsection{Handwritten Digit Classification}

\begin{figure}[t!]
\begin{center}
\includegraphics[width=2.9in]{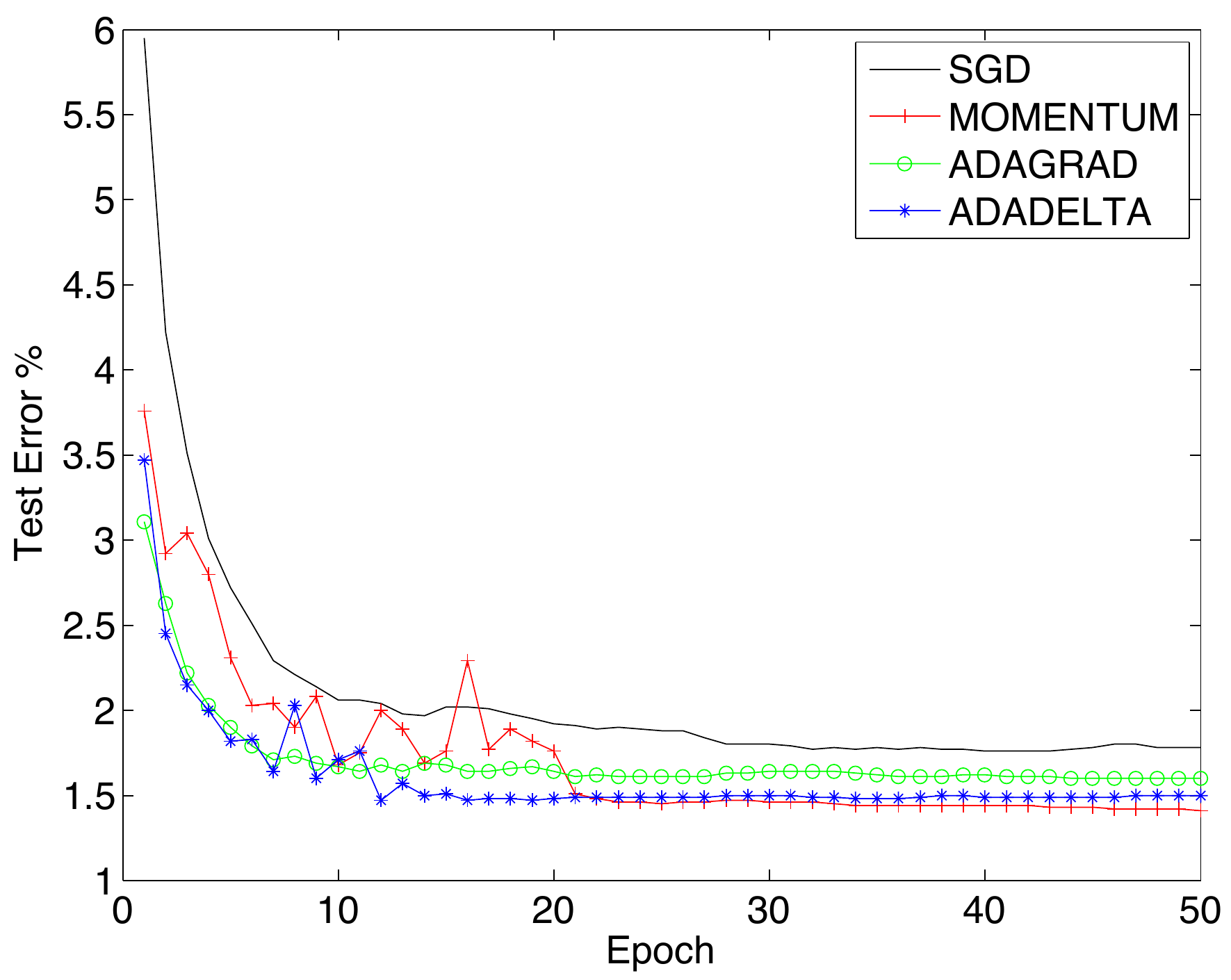}
\end{center}
\vspace*{-0.3cm}
\caption{Comparison of learning rate methods on MNIST digit classification for 50 epochs.}
\label{fig:mnist}
\vspace*{-0.3cm}
\end{figure}

In our first set of experiments we train a neural network on the MNIST handwritten digit
classification task. For comparison with Schaul \etal's method we trained with tanh nonlinearities
and 500 hidden units in the first layer followed by 300 hidden units in the second layer, with the
final softmax output layer on top.
Our method was trained on mini-batches of 100 images per batch for 6 epochs through the training
set. Setting the hyperparameters to $\epsilon = 1e-6$ and $\rho = 0.95$
we achieve $2.00\%$ test set error compared to the $2.10\%$ of Schaul \etal While this is nowhere
near convergence it gives a sense of how quickly the algorithms can optimize the classification
objective.

To further analyze various methods to convergence, we train the same neural network with 500 hidden
units in the first layer, 300 hidden units in the second layer and rectified linear activation
functions in both layers for 50 epochs.
We notice that rectified linear units work better in practice than tanh, and their
non-saturating nature further tests each of the methods at coping with large variations of
activations and gradients.

In \fig{mnist} we compare SGD, Momentum, ADAGRAD, and ADADELTA in optimizing the test set
errors. The unaltered SGD method does the worst in this case,
whereas adding the momentum term to it significantly improves performance.
ADAGRAD performs well for the first 10 epochs of training, after which it slows
down due to the accumulations in the denominator which continually increase. ADADELTA matches the
fast initial convergence of ADAGRAD while continuing to reduce the test error, converging near the
best performance which occurs with momentum.

\begin{table}[b!]
\small
\vspace*{-0mm}
\begin{center}
\begin{tabular}{|l|c|c|c|}
  \hline
  & SGD & MOMENTUM & ADAGRAD \\
   \hline
  $\epsilon = 1e^{0}$ & $\textbf{2.26\%}$ & $89.68\%$ & $43.76\%$ \\
   \hline
  $\epsilon = 1e^{-1}$ & $2.51\%$ & $\textbf{2.03\%}$ & $2.82\%$ \\
   \hline
  $\epsilon = 1e^{-2}$ & $7.02\%$ & $2.68\%$ & $\textbf{1.79\%}$ \\
   \hline
  $\epsilon = 1e^{-3}$ & $17.01\%$ & $6.98\%$ & $5.21\%$ \\
   \hline
  $\epsilon = 1e^{-4}$ & $58.10\%$ & $16.98\%$ & $12.59\%$ \\
   \hline
\end{tabular}
\vspace*{1mm}
\caption{MNIST test error rates after 6 epochs of training for various hyperparameter settings using
  SGD,  MOMENTUM, and ADAGRAD.}
\label{tab:other_hyper}
\vspace*{-5mm}
\end{center}
\end{table}

\begin{table}[b!]
\small
\vspace*{-0mm}
\begin{center}
\begin{tabular}{|l|c|c|c|}
  \hline
  & $\rho = 0.9$ & $\rho = 0.95$ & $\rho = 0.99$ \\
   \hline
  $\epsilon = 1e^{-2}$ & $2.59\%$ & $2.58\%$ & $2.32\%$ \\
   \hline
  $\epsilon = 1e^{-4}$ & $2.05\%$ & $1.99\%$ & $2.28\%$ \\
   \hline
  $\epsilon = 1e^{-6}$ & $1.90\%$ & $\textbf{1.83\%}$ & $2.05\%$ \\
   \hline
  $\epsilon = 1e^{-8}$ & $2.29\%$ & $2.13\%$ & $2.00\%$ \\
   \hline
\end{tabular}
\vspace*{1mm}
\caption{MNIST test error rate after 6 epochs for various hyperparameter settings using ADADELTA.}
\label{tab:adadelta_hyper}
\vspace*{-5mm}
\end{center}
\end{table}

\subsection{Sensitivity to Hyperparameters}

While momentum converged to a better final solution than ADADELTA after many epochs of training,
it was very sensitive to the learning rate selection, as was SGD and ADAGRAD. In
\tab{other_hyper} we vary the learning rates for each method and show the test set errors after 6
epochs of training using rectified linear units as the activation function.
The optimal settings from each column were used to generate \fig{mnist}. With
SGD, Momentum, or ADAGRAD the learning rate needs to be set to the correct order of magnitude, above
which the solutions typically diverge and below which the optimization proceeds slowly. We can
see that these results are highly variable for each method, compared to ADADELTA in
\tab{adadelta_hyper} in which the two hyperparameters do not significantly alter performance.

\subsection{Effective Learning Rates}

\begin{figure}[b!]
\vspace*{-0.3cm}
\begin{center}
\includegraphics[width=3.3in]{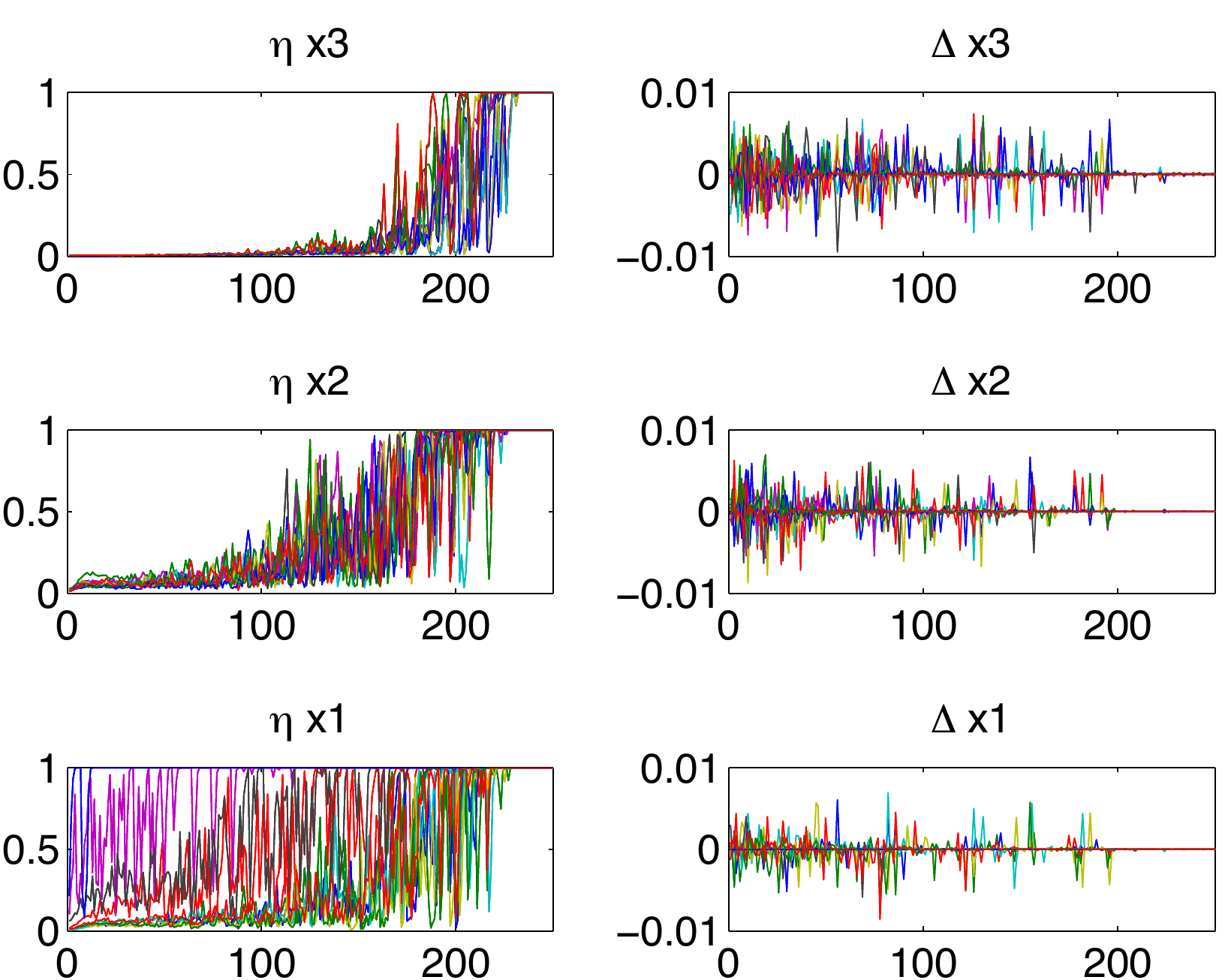}
\end{center}
\vspace*{-0.3cm}
\caption{
Step sizes and parameter updates shown every 60 batches during training the MNIST network
with tanh nonlinearities for 25 epochs.
Left: Step sizes for 10 randomly selected dimensions of each of the 3 weight matrices
of the network. Right: Parameters changes for the same 10 dimensions for each of the 3 weight
matrices. Note the large step sizes in lower layers that help compensate for vanishing gradients
that occur with backpropagation.}
\label{fig:steps}
\vspace*{-0.2cm}
\end{figure}

To investigate some of the properties of ADADELTA we plot in \fig{steps} the step sizes and
parameter updates of 10 randomly selected dimensions in each of the 3 weight matrices throughout training.
There are several interesting things evident in this figure.
First, the step sizes, or effective learning rates (all terms except $g_t$ from \eqn{adadelta})
shown in the left portion of the figure are larger for the lower layers of the network and much
smaller for the top layer at the beginning of training.
This property of ADADELTA helps to balance the fact that lower layers have smaller gradients due to
the diminishing gradient problem in neural networks and thus should have larger learning rates.

Secondly, near the end of training these step sizes converge to 1. This is typically a
high learning rate that would lead to divergence in most methods, however this convergence towards
1 only occurs near the end of training when the gradients and parameter updates are small. In this
scenario, the $\epsilon$ constants in the numerator and denominator dominate the past gradients and
parameter updates, converging to the learning rate of 1.

This leads to the last interesting property of ADADELTA which is that when the step sizes become 1,
the parameter updates (shown on the right of \fig{steps}) tend towards zero.
This occurs smoothly for each of the weight matrices effectively operating as if an annealing
schedule was present.

However, having no explicit annealing schedule imposed on the learning rate could be why momentum
with the proper hyperparameters outperforms ADADELTA later in training as seen in \fig{mnist}.
With momentum, oscillations that can occur near a minima are smoothed out,
whereas with ADADELTA these can accumulate in the numerator. An annealing schedule could possibly be
added to the ADADELTA method to counteract this in future work.

\vspace{-2mm}
\subsection{Speech Data}
In the next set of experiments we trained a large-scale neural network with 4 hidden layers on several
hundred hours of US English data collected using Voice Search, Voice IME, and read data. The network
was trained using the distributed system of \cite{gbrain-nips12} in which a centralized parameter
server accumulates the gradient information reported back from several replicas of the neural
network. In our experiments we used either 100 or 200 such replica networks
to test the performance of ADADELTA in a highly distributed environment.

The neural network is setup as in \cite{navdeep12} where the inputs are 26 frames of audio,
each consisting of 40 log-energy filter bank outputs.
The outputs of the network were 8,000 senone labels
produced from a GMM-HMM system using forced alignment with the input frames.
Each hidden layer of the neural network had 2560 hidden units and was trained with either logistic
or rectified linear nonlinearities.

\begin{figure}[t!]
\begin{center}
\includegraphics[width=2.9in]{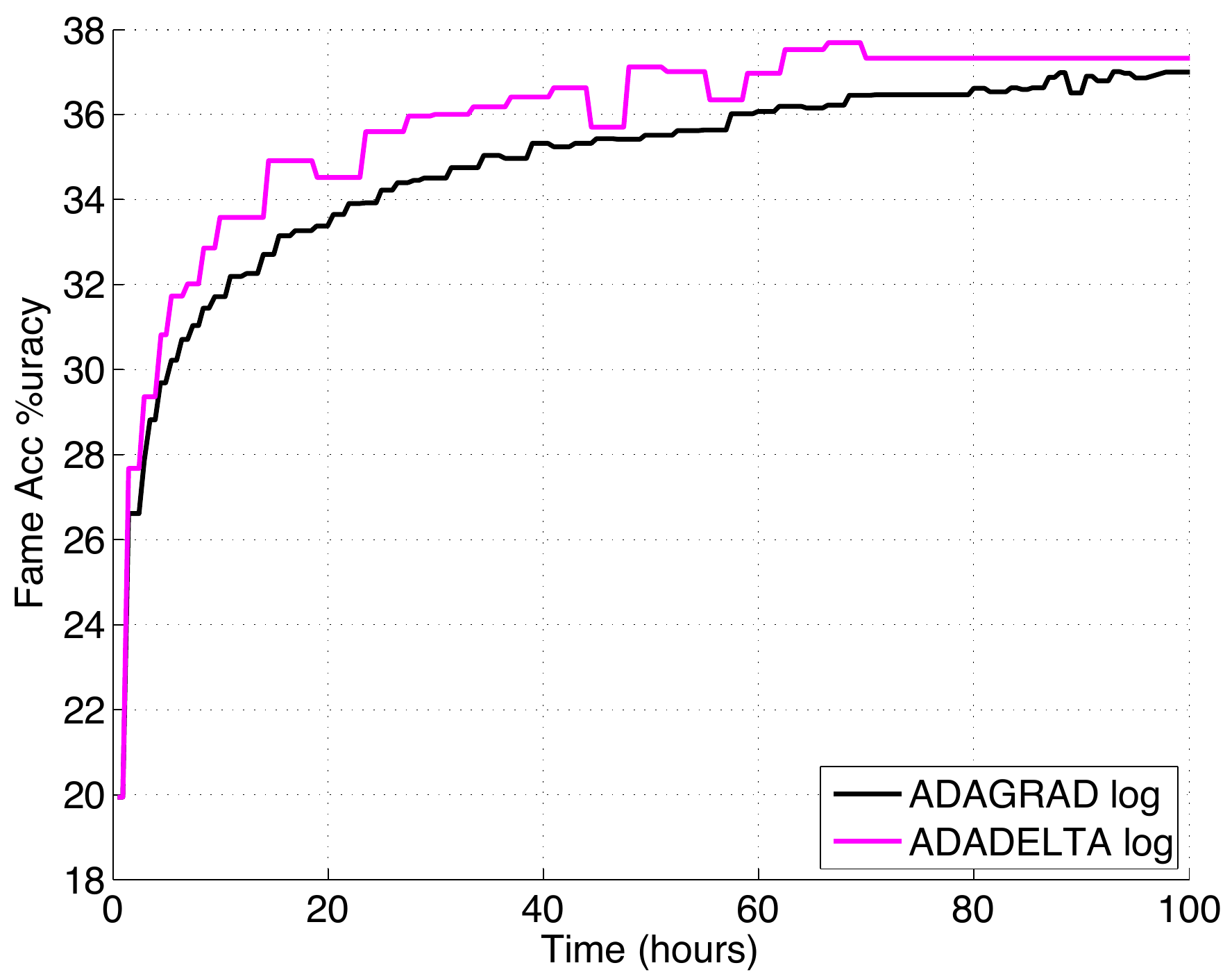}
\end{center}
\vspace*{-0.3cm}
\caption{Comparison of ADAGRAD and ADADELTA on the Speech Dataset with 100 replicas using logistic
  nonlinearities.}
\label{fig:speech_100}
\vspace*{0.3cm}
\end{figure}

\fig{speech_100} shows the performance of the ADADELTA method when using 100 network replicas.
Notice our method initially converges faster and outperforms ADAGRAD throughout training in terms of
frame classification accuracy on the test set. The same settings of $\epsilon = 1e^{-6}$ and $\rho =
0.95$ from the MNIST experiments were used for this setup.

\begin{figure}[t!]
\begin{center}
\includegraphics[width=2.9in]{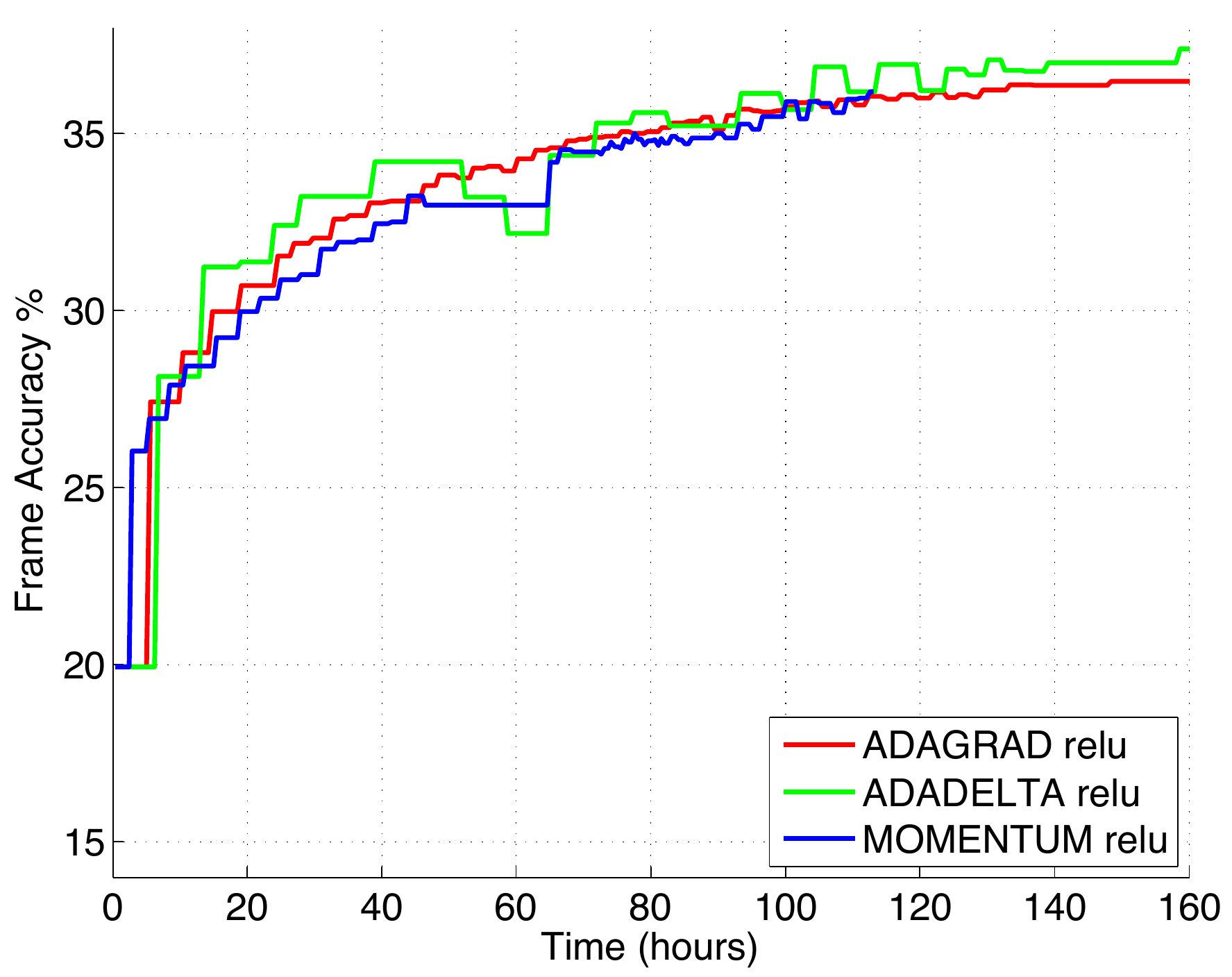}
\end{center}
\vspace*{-0.3cm}
\caption{Comparison of ADAGRAD, Momentum, and ADADELTA on the Speech Dataset with 200 replicas using
  rectified linear nonlinearities.}
\label{fig:speech_200}
\vspace*{-0.3cm}
\end{figure}

When training with rectified linear units and using 200 model replicas we also used the same
settings of hyperparameters (see \fig{speech_200}). Despite having
200 replicates which inherently introduces significants amount of noise to the gradient
accumulations, the ADADELTA method performs well, quickly converging to the same frame accuracy as
the other methods.

\label{sec:vis}

\label{sec:vis}

\section{Conclusion} \label{sec:conclusion}
In this tech report we introduced a new learning rate method based on only first order
information which shows promising result on MNIST and a large scale Speech recognition dataset. This
method has trivial computational overhead compared to SGD while providing a per-dimension learning
rate. Despite the wide variation of input data types, number of hidden units, nonlinearities and
number of distributed replicas, the hyperparameters did not need to be tuned, showing that ADADELTA
is a robust learning rate method that can be applied in a variety of situations.

\vspace{5mm}
\textbf{Acknowledgements} We thank Geoff Hinton, Yoram Singer, Ke Yang, Marc'Aurelio Ranzato and
Jeff Dean for the helpful comments and discussions regarding this work.

\bibliographystyle{IEEEbib}
\bibliography{ranzato_references}

\begin{thebibliography}{1}

\bibitem{robbins51}
H.~Robinds and S.~Monro,
\newblock ``A stochastic approximation method,''
\newblock {\em Annals of Mathematical Statistics}, vol. 22, pp. 400--407, 1951.

\bibitem{autoencoder-nn}
D.E. Rumelhart, G.E. Hinton, and R.J. Williams,
\newblock ``Learning representations by back-propagating errors,''
\newblock {\em Nature}, vol. 323, pp. 533--536, 1986.

\bibitem{adagrad}
J.~Duchi, E.~Hazan, and Y.~Singer,
\newblock ``Adaptive subgradient methods for online leaning and stochastic
  optimization,''
\newblock in {\em COLT}, 2010.

\bibitem{gbrain-nips12}
J.~Dean, G.~Corrado, R.~Monga, K.~Chen, M.~Devin, Q.~Le, M.~Mao, M.~Ranzato,
  A.~Senior, P.~Tucker, K.~Yang, and A.~Ng,
\newblock ``Large scale distributed deep networks,''
\newblock in {\em NIPS}, 2012.

\bibitem{becker1988}
S.~Becker and Y.~LeCun,
\newblock ``Improving the convergence of back-propagation learning with second
  order methods,''
\newblock Tech. {R}ep., Department of Computer Science, University of Toronto,
  Toronto, ON, Canada, 1988.

\bibitem{schaul2012}
T.~Schaul, S.~Zhang, and Y.~LeCun,
\newblock ``No more pesky learning rates,'' arXiv:1206.1106, 2012.

\bibitem{navdeep12}
N.~Jaitly, P.~Nguyen, A.~Senior, and V.~Vanhoucke,
\newblock ``Application of pretrained deep neural networks to large vocabulary
  speech recognition,''
\newblock in {\em Interspeech}, 2012.

\end{thebibliography}

\end{document}